\documentclass[12pt, english]{article}

\usepackage{sbc-template}
\usepackage{graphicx,url}
\usepackage{hyperref}

\usepackage[english]{babel}   
\usepackage[utf8]{inputenc}

\title{Low-cost Autonomous Navigation System Based on Optical Flow Classification}

\author{Michel C. Meneses\inst{1}, Leonardo N. Matos\inst{1}, Bruno O. Prado\inst{1}}

\address{Department of Computer Science -- Federal University of Sergipe
	(UFS)\\
	São Cristóvão -- SE -- Brazil
	\email{michel.meneses@dcomp.ufs.br, leonardo@dcomp.ufs.br,
		bruno@dcomp.ufs.br}
}

\begin{document}

\maketitle

\begin{abstract}
  This work presents a low-cost robot, controlled by a Raspberry Pi, whose navigation system is based on vision. The strategy used consisted of identifying obstacles via optical flow pattern recognition. Its estimation was done using the Lucas-Kanade algorithm, which can be executed by the Raspberry Pi without harming its performance. Finally, an SVM-based classifier was used to identify patterns of this signal associated with obstacles movement. The developed system was evaluated considering its execution over an optical flow pattern dataset extracted from a real navigation environment. In the end, it was verified that the acquisition cost of the system was inferior to that presented by most of the cited works, while its performance was similar to theirs.
\end{abstract}

\section{Introduction}

The development of robots capable of locomoting in an autonomous manner can be considered one of the most promising research topics in robotics \cite{Bekey2005}. Such importance is due to the almost unlimited amount of relevant and innovative applications provided by the use of those instruments \cite{Kim2012,Wang2015,Kadir2015,Chaiyasoonthorn2015}. In general, its navigation consists of four subprocesses: modeling of the environment where the robot is inserted in; localization of the robot inside that model; planning of the path to be followed by the robot; control of the robot's actuators in order to ensure the accomplishment of the planned path \cite{Siegwart2004}.

In contexts where the robot must present full autonomy, one can notice that the problem of perception and modeling of the environment becomes significantly relevant for the other subprocesses related to the navigation. This autonomy does not only refer to the complete independence of human intervention during the navigation but also to the non-utilization of any external auxiliary system species (such as radars or geolocation systems). In fact, in these situations, the only source of information about the environment used by the other subprocesses corresponds to the model built by the robot. Consequently, the detail level of this model defines the complexity and the limitations involved in the other navigation subprocesses. Among the principal sensing methods, there are those based on vision.

Sensing through vision uses cameras to obtain both dimensional and visual information, such as textures, color, luminosity, etc. In many situations these features present vital importance to the correct modeling of the environment. Moreover, monocular vision techniques demand low computational cost, especially when compared to those based on stereo vision. That is why this kind of sensing is appropriate for small, low-cost and fast robotic systems. These systems can be implemented, for instance, through a Raspberry Pi \cite{Raspberry}, which presents hardware and software specifications that are suitable for the context of autonomous navigation.

Therefore, the principal objective of this work is to develop a robotic system which is capable of locomoting in an autonomous manner through monocular vision and which is based on the low-cost Raspberry Pi platform. This paper is organized as follows: \autoref{cap:trabalhos_relacionados} consists of a bibliographic review related to autonomous navigation systems based on monocular vision; \autoref{cap:fluxo_optico} discusses the obstacle detection based on optical flow recognition; \autoref{cap:plataforma_proposta} presents the details of the navigation system proposed by this work; \autoref{cap:experimentos_e_resultados} describes the methodology used to evaluate the developed system and discusses the obtained results.

\section{Related Work}
\label{cap:trabalhos_relacionados}

Multiple autonomous navigation strategies based on monocular vision are proposed in recent publications. \autoref{tab:analise_trabalhos_relacionados} presents the main parameters of the navigation systems suggested in some of those works. It is important to mention that since each of those papers used its own evaluation methodology, the accuracy and processing rate should not be individually taken into account to establish superiority relations. By analyzing them, it becomes clear that those techniques based on floor detection demand platforms which present higher processing power. On the other hand, those techniques which considered optical flow as their input signal presented shorter processing time and lower acquisition cost. Then, the use of this signal seems more suitable for systems which have the same low-cost requirements as that intended by this work.

\begin{table}[ht]
	\centering
	\caption{Comparative analysis of the principal related works.\label{tab:analise_trabalhos_relacionados}}
	\begin{tabular}{|c|c|c|c|}
		\hline
		Strategy & Accuracy & FPS & \begin{tabular}[c]{@{}c@{}}Cost of the \\ Platform (US\$)\end{tabular} \\ \hline
		\begin{tabular}[c]{@{}c@{}}Floor detection \\ by homography \\ \cite{Conrad2010}\end{tabular} & 99,60\% & - & 7.142,82 \\ \hline
		\begin{tabular}[c]{@{}c@{}}Floor detection by \\ line segmentation \\ \cite{Li2010}\end{tabular} & 89,10\% & 5 & 10.612,82 \\ \hline
		\begin{tabular}[c]{@{}c@{}}Optical flow \\ segmentation \\ \cite{Caldeira2007}\end{tabular} & - & 7,41 & 4.080,00 \\ \hline
		\begin{tabular}[c]{@{}c@{}}Optical flow \\ classification \\ \cite{Shankar2014}\end{tabular} & 88,80\% & 7 & 50,00\\
		\hline
	\end{tabular}
\end{table}

\section{Obstacle Detection by Optical Flow Classification}
\label{cap:fluxo_optico}

\begin{figure}[htb]
	\centering
	\begin{minipage}[t]{0.4\linewidth}
		\includegraphics[width=\linewidth]{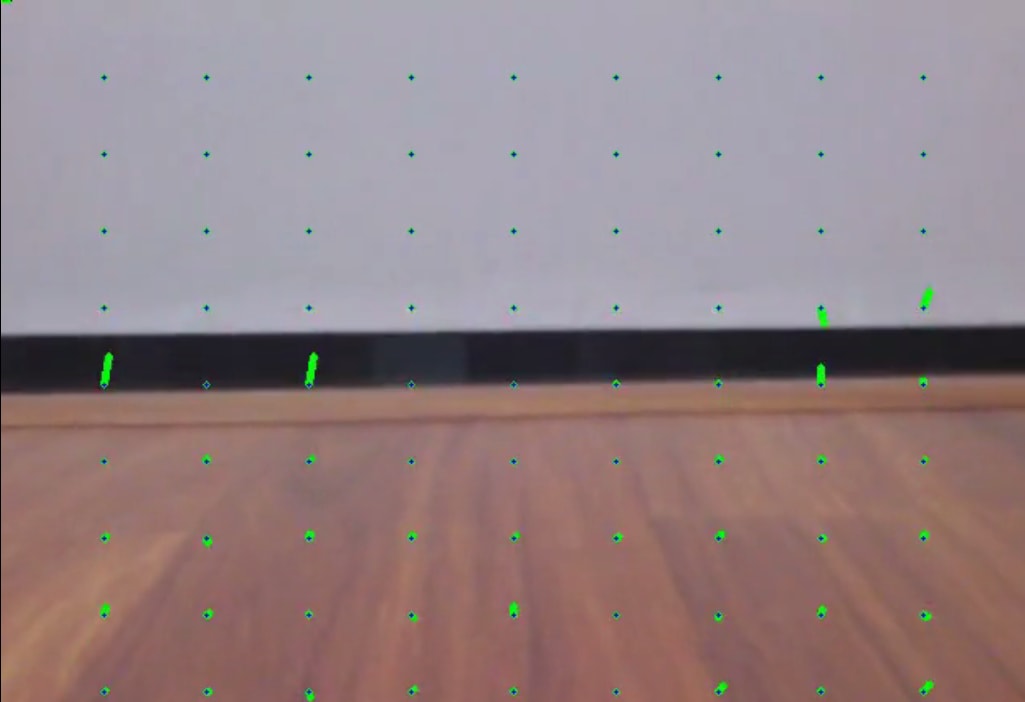}
		\caption{Optical flow from a scene without near obstacles.\label{fig:movimentacao_sem_obstaculo}}
	\end{minipage}
	\begin{minipage}[t]{0.4\linewidth}
		\includegraphics[width=\linewidth]{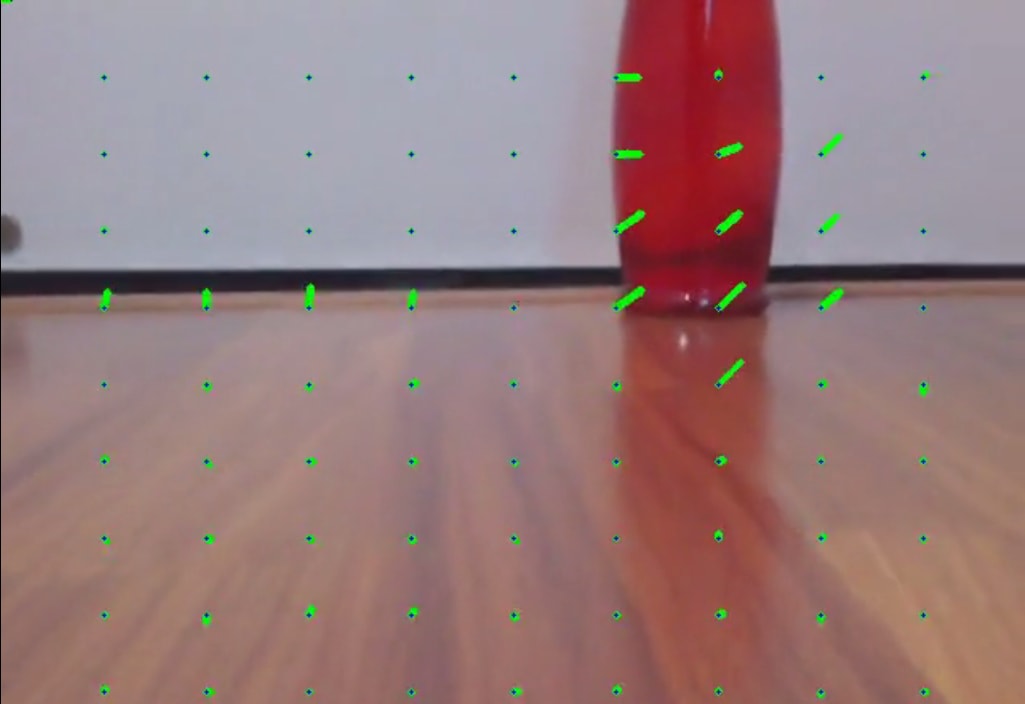}
		\caption{Contrasting motion of an obstacle with respect to the remaining elements in the environment.\label{fig:movimentacao_com_obstaculo}}
	\end{minipage}
\end{figure}

Let $t$, defined by the function $I(\vec{x}, t)$, be the brightness intensity of an image, where $\vec{x} = (x, y)^T$ corresponds to the position of each pixel. Considering that at $t+1$ such intensity is translated and holds, it follows that:

\begin{equation}\label{eq:constancia_brilho}
I(\vec{x}, t) = I(\vec{x} + \vec{u}, t+1),
\end{equation}

\noindent where $\vec{u} = (u_1, u_2)^T$ denotes the displacement on the 2D plan. Such a vector equals to the optical flow, which describes the apparent motion of an image intensity pattern. This motion is generally associated with the relative displacement between objects included in an image sequence and the camera. Consequently, it is possible to measure such movement through this flow \cite{Horn1981}. For the problem of autonomous navigation, this kind of information is extremely relevant since it is directly related to the motion realized by the robot about the other elements included in the environment. Among the principal applications of this knowledge, there is the obstacle detection. Obstacles have as central feature their accent relative motion. Such movement is contrasting to that presented by the remaining elements in the scene, once these are further from the agent (\autoref{fig:movimentacao_sem_obstaculo} and \autoref{fig:movimentacao_com_obstaculo}).

Therefore, it follows that by optical flow samples extracted from many scenes with and without obstacles it is possible to build a model which is capable of separating such patterns. In other words, it is possible to train a classifier from motion patterns already labeled and to apply it to indicate if a new scene presents or not an obstacle. Thus, obstacle detection based on optical flow recognition can be performed by a classifier machine, whose supervised learning is achieved based on the following definitions:

\begin{itemize}
	\item $\vec{x_i}=(F_1,...F_n)$ corresponds to the feature vector associated with the $i$-th image in the considered sequence;
	\item $F_n$ corresponds to the pair $(v_1, v_2)$, where $v_1$ and $v_2$ are the amplitude and the phase of the optical flow vector associated with the $n$-th image point, respectively;
	\item Each $\vec{x_i}$ is related to a value $y_i\in\{-1,+1\}$, where the labels $-1$ and $+1$ indicate the absence and the presence of an obstacle in the image, respectively.
\end{itemize}

\section{Developed System}
\label{cap:plataforma_proposta}

The navigation system developed in this work is based on obstacle detection by optical flow classification. The whole system was built on the low-cost Raspberry Pi platform. The following sections describe both hardware and software of the developed robot.

\subsection{Hardware}
\begin{figure}[htb]
	\centering
	\includegraphics[width=0.7\linewidth]{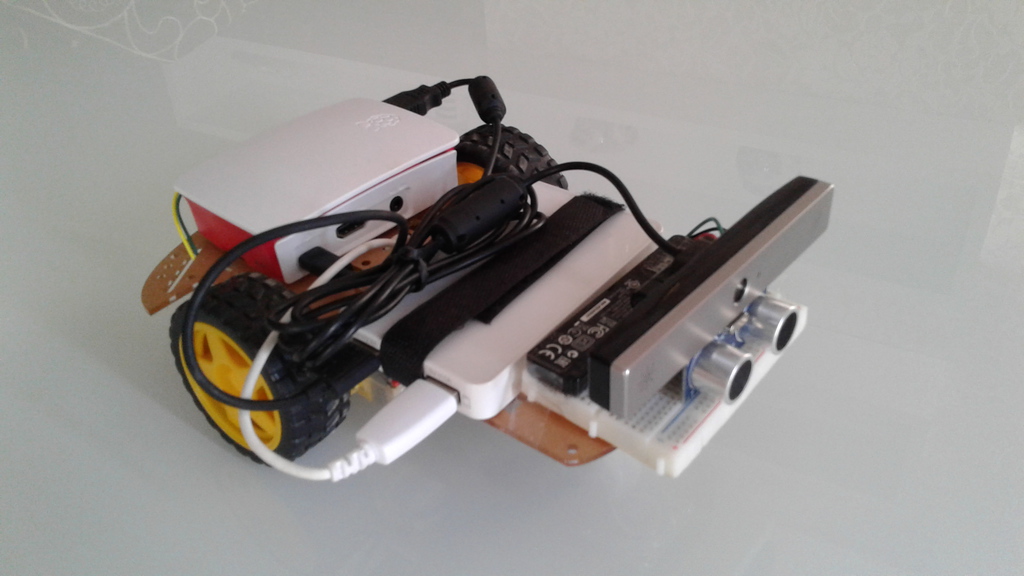}
	\caption{Perspective view of the built robotics platform. \label{fig:plataforma_construida}}
\end{figure}

The components that constitute the hardware of the designed platform correspond to a Raspberry Pi computer, a sustain chassis which contains two actuators, a power supply, and two sensors: a monocular camera and an ultrasonic sensor. \autoref{fig:plataforma_construida} presents the final built platform. Its financial cost is described in \autoref{tab:custo_plataforma}.

\begin{table}[h]\caption{Cost of the developed platform.\label{tab:custo_plataforma}}
		\centering
		\begin{tabular}{|c|c|}
			\hline
			Component & Cost (US\$) \\ \hline
			Raspberry Pi 3 & 30,00 \\ \hline
			LG AN-VC500 Camera & 89,99 \\ \hline
			Chassis & 21,55 \\ \hline
			HC-SR04 Sensor & 2,00 \\ \hline
			L293D & 1,90 \\ \hline
			Power Bank APC M5WH & 25,38 \\ \hline
			\textbf{Total} & 170,82 \\
			\hline
		\end{tabular}
\end{table}

\subsection{Software}
\begin{figure}[htb]
	\centerline{\includegraphics[width=0.9\linewidth]{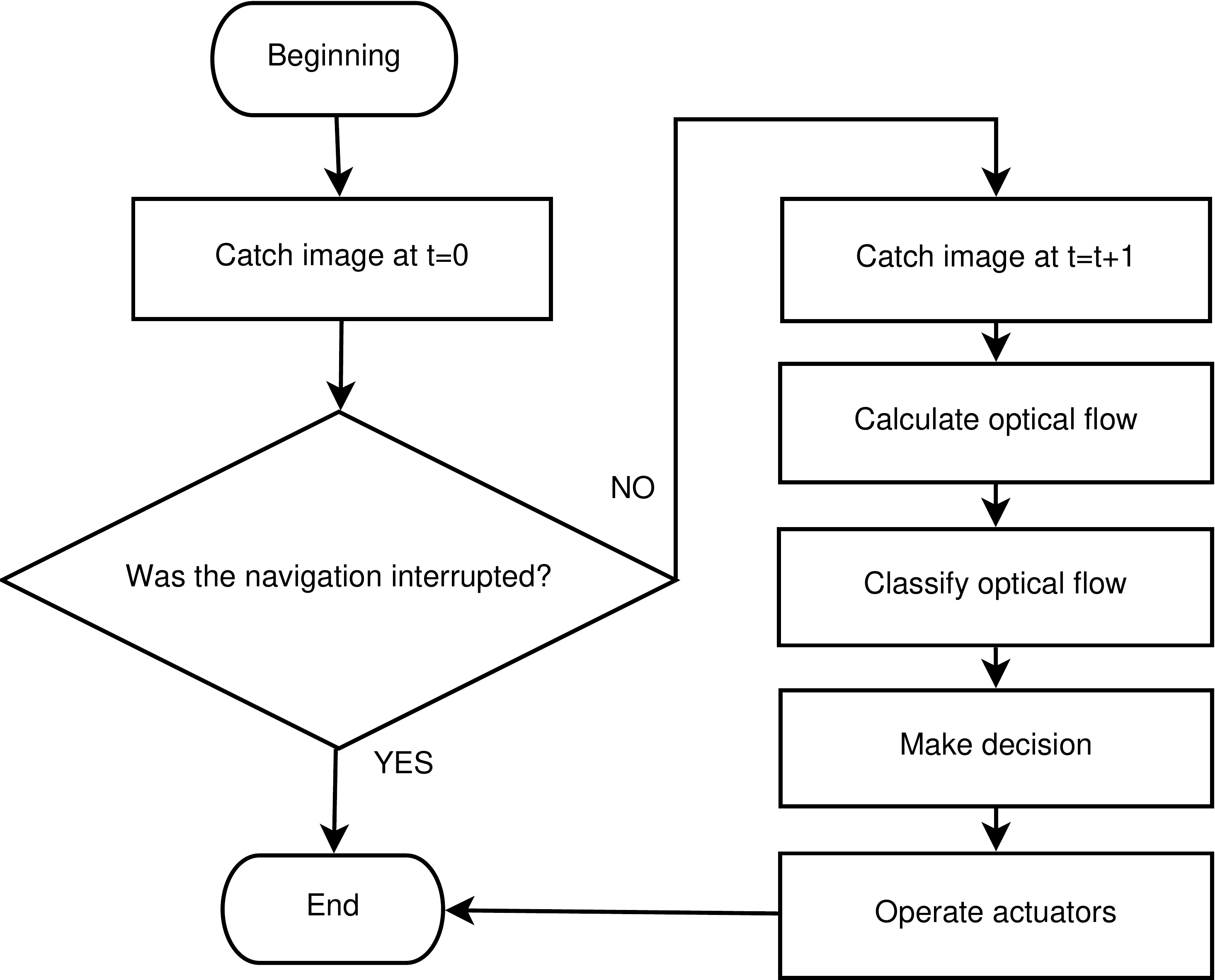}}
	\caption{Flowchart of the proposed navigation system.\label{fig:fluxograma_sistema}}
\end{figure}

The proposed navigation system was developed based on the computer vision and machine learning libraries OpenCV \cite{OpenCV3-2} and Scikit-Learn \cite{scikit-learn}, respectively. Its work cycle is shortly described in \autoref{fig:fluxograma_sistema}. According to that flowchart, the system initially catches a referential image and starts the navigation cycle. Then, another image is acquired and the optical flow from those two images is estimated. This flow is presented to an SVM classifier \cite{Chih-WeiHsuChih-ChungChang2008} with RBF kernel, which indicates if there is an obstacle on the path to be followed. Based on that indication, a decision related to the update of the robot's direction is made. In the cases where obstacles are detected, a deflection to the direction with lower optical flow intensity is made, since this direction presents less relative motion and, hence, has a lower probability of containing new obstacles.

\begin{figure}[htb]
	\centering
	\begin{minipage}[t]{0.4\linewidth}
		\includegraphics[width=\linewidth]{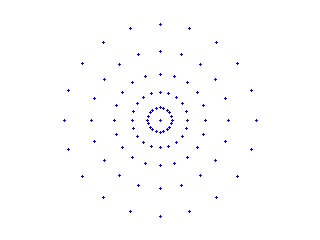}
		\caption{Points distribution considered to estimate the optical flow.\label{fig:distribuicao_pontos_monitorados}}
	\end{minipage}
	\qquad
	\begin{minipage}[t]{0.4\linewidth}
		\includegraphics[width=\linewidth]{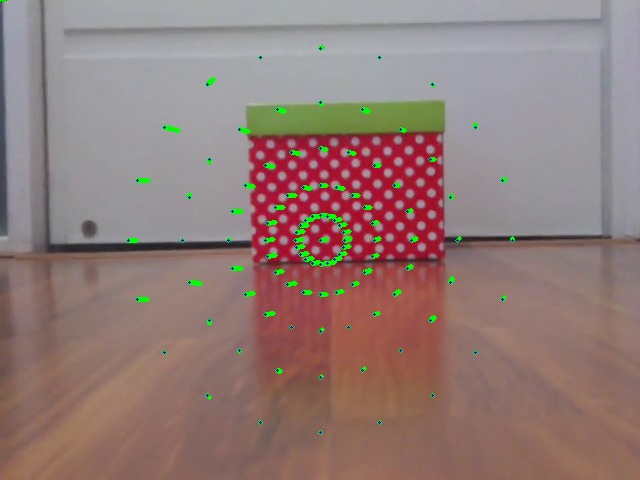}
		\caption{Optical flow estimated during the navigation of the robot.\label{fig:imagem_fluxo_optico}}
	\end{minipage}
\end{figure}

The Lucas-Kanade algorithm \cite{Lucas1981} was used to estimate the optical flow from two environment images taken in a row. In order to apply it, a circular symmetrical observation points distribution was considered (\autoref{fig:distribuicao_pontos_monitorados}). This distribution consists of 1 central point surrounded by 5 concentric rings, each one formed by 20 equally spaced points. The distance between each of these rings and the central point increases exponentially. \autoref{fig:imagem_fluxo_optico} illustrates the flow estimated from a scene registered during the robot's navigation.

\section{System Evaluation}
\label{cap:experimentos_e_resultados}

\subsection{Off-line Evaluation}
\label{sec:avaliacao_offline}
The off-line evaluation considered only the quality of the SVM classifier. The following sections describe its details.

\subsubsection{Methodology}
The off-line methodology consisted of the k-fold cross validation \cite{kohavi1995study} where $k = 8$. The correct and incorrect indications of the classifier during all the process were registered for both classes, in order to fill out its confusion matrix. By using this, it was possible to extract measures which clearly express the quality of the classifier's performance. Such measures correspond to the precision, the recall, the F measure and the accuracy.

\subsubsection{Dataset}
\begin{figure}[htb]
	\centerline{\includegraphics[width=0.7\linewidth]{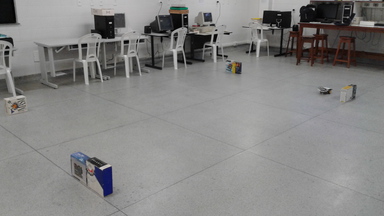}}
	\caption{Environment where the dataset videos were recorded.\label{fig:ambiente_teste}}
\end{figure}

The dataset used during the classifier's validation process was built in this work since no repositories containing optical flow patterns labeled according to the presence of obstacles were found. Therefore, that dataset was built through optical flow patterns extracted from 8 videos whose resolution equals to $320\times240$ pixels. Those videos were recorded by the robotic platform developed in this work. In order to accomplish this, the robot was remotely guided by a human controller throughout a circuit which contained different obstacles. Moreover, the features of the environment were changed as long as new videos were recorded. Finally, the labels of the built dataset were created by measures obtained through the ultrasonic sensor HC-SR04, which was already set up on the platform. \autoref{fig:ambiente_teste} shows the environment run by the robot.

\subsubsection{Baseline}
The same methodology which was used to evaluate the SVM classifier was also applied to other two learning models: a Perceptron \cite{Rosenblatt1958} and an SVR classifier (Support Vector Regressor) with RBF kernel \cite{Drucker1997}. The first one consisted of a linear classification model whose training parameters corresponded to a maximum limit of 100 iterations and balanced weight of examples of each class. On the other hand, the second model consisted of a regression model, through which it is possible to estimate the distance associated with each flow pattern.

\subsubsection{Results}
\autoref{tab:matriz_confusao_baselines} presents both the performance measures extracted from the SVM classifier and those associated with the baseline models. One can notice that the SVM classifier's recall value was relatively low. However, in a real situation, it is only necessary to recognize one flow sample as related to an obstacle in order to apply an avoidance maneuver. In other words, the recall value does not necessarily equal the collision rate. On the other hand, the classifier's precision is more relevant since the lower this measure is the higher the number of mistaken deflections tends to be. By analyzing the mean measured precision it is possible to notice that its value is relatively superior to the recall, which was expected.

\begin{table}[htb]
	\centering
	\caption{Mean measures assessed based on the different models considered.\label{tab:matriz_confusao_baselines}}
		\begin{tabular}{|c|c|c|c|c|}
			\hline
			Model & Precision & Recall & F Measure & Accuracy \\ \hline
			SVM & $75,46\pm6,21\%$  & $61,71\pm4,75\%$ & $68,00\pm3,75\%$ & $89,90\pm1,36\%$\\ \hline
			Perceptron & $16,37\pm2,36\%$  & $45,92\pm9,89\%$ & $24,08\pm3,81\%$ & $50,83\pm3,09\%$\\ \hline
			SVR & -  & $0,00\pm0,00\%$ & - & $82,84\pm1,08\%$\\
			\hline
		\end{tabular}
\end{table}

Nevertheless, the processing frequency presented by the system, without considering the time to catch the images, was equal to 14,88FPS. When considering the average system's capture frequency as equal to 25,28FPS (assessed experimentally), it can be concluded that its average full processing rate was equal to 9,4FPS.

\autoref{tab:analise_sistema_vs_trabalhos_relacionados} shows a final comparison between the developed system and those suggested by those works cited in \autoref{cap:trabalhos_relacionados}. Based on the comparative parameters presented, it can be concluded that the developed navigation system has a low cost, a high processing frequency, and a satisfactory accuracy.

\begin{table}[h]
	\centering
	\caption{Comparative analysis between the developed system and those suggested by the works cited in \autoref{cap:trabalhos_relacionados}.\label{tab:analise_sistema_vs_trabalhos_relacionados}}
		\begin{tabular}{|c|c|c|c|}
			\hline
			System & Accuracy & \begin{tabular}[c]{@{}c@{}}Frames \\ Per Second\end{tabular} & \begin{tabular}[c]{@{}c@{}}Cost of the \\ Platform (US\$)\end{tabular}\\ \hline
			\begin{tabular}[c]{@{}c@{}}Floor detection by homography \\\cite{Conrad2010}\end{tabular} & 99,60\% & - & 7142,82 \\ \hline
			\begin{tabular}[c]{@{}c@{}}Floor detection by line segmentation \\\cite{Li2010}\end{tabular} & 89,10\% & 5 & 10612,82 \\ \hline
			\begin{tabular}[c]{@{}c@{}}Optical flow segmentation \\\cite{Caldeira2007}\end{tabular} & - & 7,41 & 4080,00 \\ \hline
			\begin{tabular}[c]{@{}c@{}}Optical flow classification \\\cite{Shankar2014}\end{tabular} & 88,80\% & 7 & 50\\ \hline
			\begin{tabular}[c]{@{}c@{}}\textbf{Developed}\end{tabular} & \textbf{89,90\%} & \textbf{9,4} & \textbf{170,82} \\
			\hline
		\end{tabular}
\end{table}

\subsection{On-line Evaluation}
\begin{figure}[htb]
	\centering
	\begin{minipage}[t]{0.338\linewidth}
		\includegraphics[width=\linewidth]{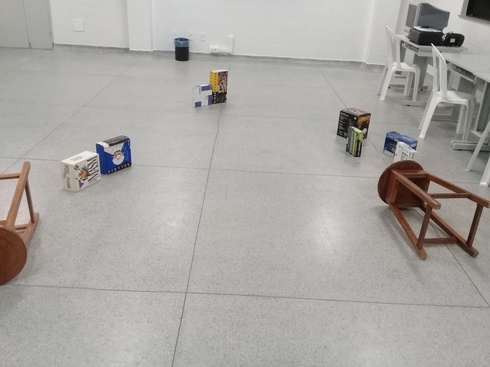}
		\caption{Navigation environment used during the on-line evaluation.\label{fig:ambiente_teste_online}}
	\end{minipage}
	\qquad
	\begin{minipage}[t]{0.452\linewidth}
		\includegraphics[width=\linewidth]{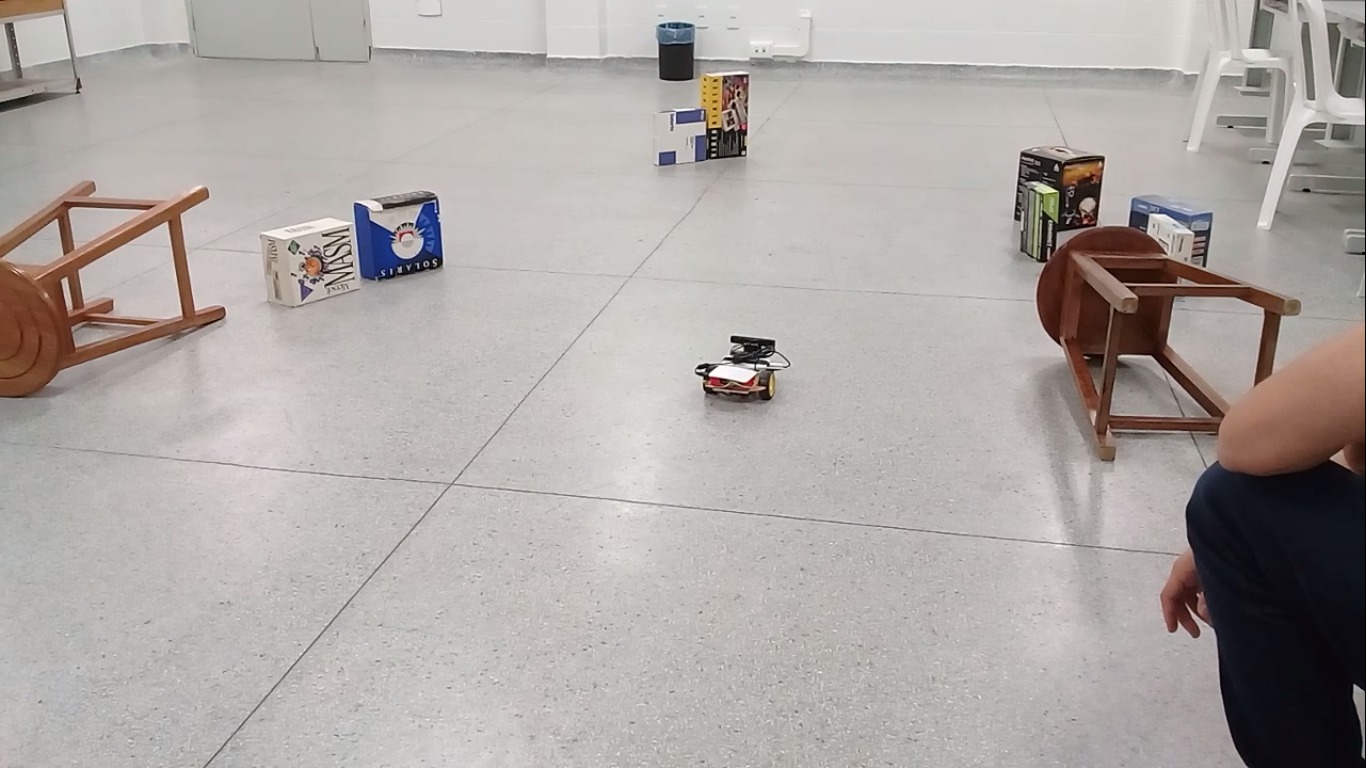}
		\caption{Robot's autonomous navigation throughout the test circuit. \label{fig:avaliacao_online}}
	\end{minipage}
\end{figure}

The on-line evaluation consisted of inserting the robot into a specific environment and watching its autonomous navigation. Then, the circuit showed in \autoref{fig:ambiente_teste_online} was considered. In order to allow the robot to recognize obstacles, the SVM trained during the off-line experiment (\autoref{sec:avaliacao_offline}) was embedded into the platform. Finally, it is import to highlight that some of the obstacles present in the on-line test circuit were not used to build the classifier training dataset.

The video provided at \url{http://www.youtube.com/v/hzyKAGhQExg?rel=0} shows the robot's navigation throughout the test circuit (\autoref{fig:avaliacao_online}). By analyzing it, one can highlight that at no moment the robot had collided with any of the obstacles inserted into the circuit. Moreover, it can be observed that every avoidance maneuver was made to the right direction. Finally, it is realized that the actuators control allowed the precise execution of the maneuvers, through which the robot could place itself in the correct direction to the subsequent obstacles.

\section{Conclusion}
\label{cap:conclusao}
This work had explored the definition of optical flow and its application in obstacle recognition. Based on the performed investigation, an autonomous robot capable of identifying obstacles by optical flow classification was developed. This navigation system was based on the Lucas-Kanade algorithm and in an SVM classifier. This system had been evaluated based on its off-line and on-line performance. In the end, it could be verified that the built system had exhibited a higher accuracy and a lower cost than the majority of the cited works. Also, its processing frequency had overcome those shown by the related works. Finally, it was possible to prove the successful behavior of the whole system in a real navigation circuit.

For future works, one could exploit the use of probability classification models by which would be possible to develop a navigation system capable of predicting obstacles based on optical flow patterns estimated in the past \cite{Lipton2015}. Also, one could experiment new strategies to extract the features of the optical flow assessed from a scene, in a way that the number of dimensions analyzed by the classifier would be reduced.

\bibliographystyle{sbc}
\bibliography{bibliografia}

\begin{thebibliography}{}

\bibitem[Bekey 2005]{Bekey2005}
Bekey, G.~A. (2005).
\newblock {\em Autonomous Robots: From Biological Inspiration to Implementation
  and Control (Intelligent Robotics and Autonomous Agents series)}.
\newblock A Bradford Book.

\bibitem[Caldeira et~al. 2007]{Caldeira2007}
Caldeira, E. M. D.~O., Schneebeli, H. J.~A., and Sarcinelli-Filho, M. (2007).
\newblock {An optical flow-based sensing system for reactive mobile robot
  navigation}.
\newblock {\em Sba: Controle {\&} Automa{\c{c}}{\~{a}}o Sociedade Brasileira de
  Automatica}, 18(3):265--277.

\bibitem[Chaiyasoonthorn et~al. 2015]{Chaiyasoonthorn2015}
Chaiyasoonthorn, S., Hongyim, N., and Mitatha, S. (2015).
\newblock Building automatic packet report system to report position and
  radiation data for autonomous robot in the disaster area.
\newblock In {\em 2015 15th International Conference on Control, Automation and
  Systems ({ICCAS})}. Institute of Electrical and Electronics Engineers
  ({IEEE}).

\bibitem[{Chih-Wei Hsu, Chih-Chung Chang} and Lin
  2008]{Chih-WeiHsuChih-ChungChang2008}
{Chih-Wei Hsu, Chih-Chung Chang} and Lin, C.-J. (2008).
\newblock {A Practical Guide to Support Vector Classification}.
\newblock {\em BJU international}, 101(1):1396--400.

\bibitem[Conrad and DeSouza 2010]{Conrad2010}
Conrad, D. and DeSouza, G.~N. (2010).
\newblock {Homography-based ground plane detection for mobile robot navigation
  using a Modified EM algorithm}.
\newblock {\em 2010 IEEE International Conference on Robotics and Automation},
  pages 910--915.

\bibitem[Drucker et~al. 1997]{Drucker1997}
Drucker, H., Burges, C. J.~C., Kaufman, L., Smola, A., and Vapnik, V. (1997).
\newblock {Support vector regression machines}.
\newblock {\em Advances in Neural Information Processing Systems}, 1:155--161.

\bibitem[Foundation 2016]{Raspberry}
Foundation, R.~P. (2016).
\newblock Raspberry pi - teach, learn, and make with raspberry pi.

\bibitem[Horn and Schunck 1981]{Horn1981}
Horn, B. K.~P. and Schunck, B.~G. (1981).
\newblock {Determining optical flow}.
\newblock {\em Artificial Intelligence}, 17(1-3):185--203.

\bibitem[Kadir et~al. 2015]{Kadir2015}
Kadir, M.~A., Chowdhury, M.~B., Rashid, J.~A., Shakil, S.~R., and Rhaman, M.~K.
  (2015).
\newblock An autonomous industrial robot for loading and unloading goods.
\newblock In {\em 2015 International Conference on Informatics, Electronics
  {\&} Vision ({ICIEV})}. Institute of Electrical and Electronics Engineers
  ({IEEE}).

\bibitem[Kim et~al. 2012]{Kim2012}
Kim, D., uk~Shin, J., Kim, H., Lee, D., Lee, S.-M., and Myung, H. (2012).
\newblock Development of jellyfish removal robot system {JEROS}.
\newblock In {\em 2012 9th International Conference on Ubiquitous Robots and
  Ambient Intelligence ({URAI})}. Institute of Electrical and Electronics
  Engineers ({IEEE}).

\bibitem[Kohavi et~al. 1995]{kohavi1995study}
Kohavi, R. et~al. (1995).
\newblock A study of cross-validation and bootstrap for accuracy estimation and
  model selection.
\newblock In {\em Ijcai}, volume~14, pages 1137--1145. Stanford, CA.

\bibitem[Li and Birchfield 2010]{Li2010}
Li, Y. and Birchfield, S.~T. (2010).
\newblock {Image-based segmentation of indoor corridor floors for a mobile
  robot}.
\newblock {\em IEEE/RSJ 2010 International Conference on Intelligent Robots and
  Systems, IROS 2010 - Conference Proceedings}, pages 837--843.

\bibitem[Lipton et~al. 2015]{Lipton2015}
Lipton, Z.~C., Berkowitz, J., and Elkan, C. (2015).
\newblock {A Critical Review of Recurrent Neural Networks for Sequence
  Learning}.
\newblock pages 1--38.

\bibitem[Lucas and Kanade 1981]{Lucas1981}
Lucas, B.~D. and Kanade, T. (1981).
\newblock {An Iterative Image Registration Technique with an Application to
  Stereo Vision}.
\newblock {\em Imaging}, 130(x):674--679.

\bibitem[Pedregosa et~al. 2011]{scikit-learn}
Pedregosa, F., Varoquaux, G., Gramfort, A., Michel, V., Thirion, B., Grisel,
  O., Blondel, M., Prettenhofer, P., Weiss, R., Dubourg, V., Vanderplas, J.,
  Passos, A., Cournapeau, D., Brucher, M., Perrot, M., and Duchesnay, E.
  (2011).
\newblock Scikit-learn: Machine learning in {P}ython.
\newblock {\em Journal of Machine Learning Research}, 12:2825--2830.

\bibitem[Rosenblatt 1958]{Rosenblatt1958}
Rosenblatt, F. (1958).
\newblock The perceptron: A probabilistic model for information storage and
  organization in the brain.
\newblock {\em Psychological Review}, 65(6):386--408.

\bibitem[Shankar et~al. 2014]{Shankar2014}
Shankar, A., Vatsa, M., and Sujit, P.~B. (2014).
\newblock {Collision avoidance for a low-cost robot using SVM-based monocular
  vision}.
\newblock {\em 2014 IEEE International Conference on Robotics and Biomimetics,
  IEEE ROBIO 2014}, pages 277--282.

\bibitem[Siegwart and Nourbakhsh 2004]{Siegwart2004}
Siegwart, R. and Nourbakhsh, I.~R. (2004).
\newblock {\em Introduction to Autonomous Mobile Robots}.
\newblock Massachusetts Institute of Technology.

\bibitem[Team 2017]{OpenCV3-2}
Team, O.~D. (2017).
\newblock Opencv 3.2 | opencv.

\bibitem[Wang et~al. 2015]{Wang2015}
Wang, L., Chen, F., and Yin, H. (2015).
\newblock {Detecting and tracking vehicles in traffic by unmanned aerial
  vehicles}.
\newblock {\em Automation in Construction}, (May).

\end{thebibliography}

\end{document}